\pdfoutput=1

\documentclass[11pt]{article}

\usepackage[]{EMNLP2022}

\usepackage{times}
\usepackage{latexsym}

\usepackage[T1]{fontenc}
\usepackage[utf8]{inputenc}

\usepackage{microtype}

\usepackage{graphicx}
\usepackage{url}
\usepackage{latexsym}
\usepackage{makecell}
\usepackage{epstopdf}
\usepackage{tipa}
\usepackage{hyperref}
\usepackage{xstring}
\usepackage{amsfonts}
\usepackage{amsmath}
\usepackage{fixltx2e}
\usepackage[T1]{fontenc}
\usepackage{booktabs}
\usepackage{arabtex}
\usepackage{hhline}
\usepackage{pdfpages}
\usepackage{utf8}
\newcommand{\hide}[1]{}

\newcommand{\AHAMZAUP}{{\^{A}}}

\newcommand{\TAMARBUTA}{{$\hbar$}}

\newcommand{\DHA}{{\dh}}
\newcommand{\SHIN}{{\v{s}}}
\newcommand{\AYN}{{$\varsigma$}}

\newcommand{\SHADDA}{{$\sim$}}

\usepackage{multirow}
\usepackage{subfigure}
\usepackage{amssymb}
\usepackage[super]{nth}
\usepackage{xspace}

\usepackage[size=tiny,color=pink]{todonotes}

\usepackage{caption}
\setcode{utf8}
\vocalize
\usepackage{xcolor}

\newcommand{\caphi}[1]{{/{{\it #1}}/}}
\newcommand{\curras}{{Curras}\xspace}
\newcommand{\maknuune}{{Maknuune}\xspace}

\title{Maknuune: A Large Open Palestinian Arabic Lexicon}

\hide{

}

\author{Shahd Dibas,\textsuperscript{\textdagger} Christian Khairallah,\textsuperscript{\textdaggerdbl}
Nizar Habash\textsuperscript{\textdaggerdbl}\\
\bf Omar Fayez Sadi,\textsuperscript{\textasteriskcentered} 
 Tariq Sairafy,\textsuperscript{\textasteriskcentered}
 Karmel Sarabta,\textsuperscript{\textasteriskcentered}
 Abrar Ardah\textsuperscript{\textasteriskcentered}
  \\
\textsuperscript{\textdagger}University of Oxford, \textsuperscript{\textdaggerdbl}New York University Abu Dhabi\\
\textsuperscript{\textasteriskcentered}University College of Educational Sciences - UNRWA\\
\small{\texttt{shahd.dibas@ling-phil.ox.ac.uk},
\texttt{christian.khairallah@nyu.edu},
\texttt{nizar.habash@nyu.edu}}}

\begin{document}
\maketitle
\begin{abstract}
We present Maknuune~\<مكنونة>, 
a large open lexicon for the Palestinian Arabic dialect. Maknuune has 
over 36K entries from 17K lemmas, and 3.7K roots. 
All entries include diacritized Arabic orthography, phonological transcription and English glosses. 
Some entries are enriched with additional information such as  broken plurals and templatic feminine forms, associated phrases and collocations, Standard Arabic glosses, and  examples or notes on grammar, usage, or location of collected entry. 
%
%
\end{list} 
\end{abstract}


\section{Introduction}
 
Arabic is a collective of historically related variants that co-exist in a diglossic \cite{Ferguson:1959:diglossia} relationship between a Standard variant and geographically specific dialectal variants. Standard Arabic (SA, \<العربية الفصحى>)
is typically used to refer to the older Classical Arabic (CA) used in Quranic texts and pre-islamic poetry, all the way to Modern SA (MSA), the official language of news and culture in the Arab World.  Dialectal Arabic (DA) is classified geographically into regions such as Egyptian, Levantine, Maghrebi, and Gulf.
The dialects, which  differ among themselves and SA, are the primary mode of spoken communication, although increasingly they are dominating in written form on social media.  That said, DA has no official prescriptive grammars or orthographic standards, unlike the highly standardized and regulated MSA.  In the realm of natural language processing (NLP), MSA has relatively more annotated and parallel resources than DA; although there are many notable efforts to fill gaps in all Arabic variants \cite{alyafeai2022masader}.

In this paper, we focus on Palestinian Arabic (PAL), which is part of the South Levantine Arabic dialect subgroup. PAL consists of several sub-dialects in the region of Historic Palestine that 
vary in terms of their phonology and lexical choice \citep{Jarrar:2016:curras}. 
PAL, like all other DA,  has been historically influenced by many languages, specifically, in its case, Syriac, Turkish, Persian, English and most recently Modern Hebrew \cite{moin2019etymological}, as well as other Arabic dialects that came in interaction with PAL after the Nakba. 
While this research effort was originally motivated by the need to document and preserve the cultural heritage and unique identities 
of the various PAL sub-dialects, it has expanded to cover PAL's ever-evolving nature as a living language, and provides a resource to support research and development in Arabic dialect NLP.

Concretely, we present \textbf{Maknuune}~\<مكنونة>,\footnote{\<مكنونة>~/maknūne/ is a PAL farming term that refers to an egg intentionally left behind in a specific location to encourage the chicken to lay more eggs in that location.
We hope that the lexicon will encourage other researchers and citizen linguists to contribute to it.}
%
a large open lexicon for PAL, with over 36K entries from 17K lemmas, and 3.7K roots.\footnote{In this initial phase of Maknuune, we focus on the PAL sub-dialects spoken in the West Bank, an area with dialectal diversity across many dimensions such as \textit{lifestyle} (urban, rural, bedouin), religion, gender, and social class.}
All entries include diacritized Arabic orthography and phonological transcription following \newcite{Habash:2018:unified}, as well as English glosses. Important inflectional variants are included for some lemmas, such as broken plural and templatic feminine. 
About 10\%  of the entries are phrases (multiword expressions) indexed by their primary lemmas. And about 67\%  
of the entries include MSA glosses,  examples, and/or notes on grammar, usage, or location of collected entry.
To our knowledge, Maknuune is the largest open machine-readable dictionary for PAL. Maknuune is publicly viewable and downloadable.\footnote{\url{www.palestine-lexicon.org}}
%

We discuss some related work in Section~\ref{related}, and highlight some PAL linguistic facts 
that motivated many of our 
design choices in Section~\ref{lingfacts}.  Section~\ref{method} presents our data collection process and annotation guidelines. We present statistics for our lexicon and evaluate its coverage 
in Section~\ref{eval}.


\section{Related Work} 
\label{related}

\paragraph{Linguistic Descriptions} There are several linguistic references describing various aspects of PAL \citep{Rice:1979:eastern, herzallah1990aspects, hopkins1995sarar, elihai2004olive, talmon200419th, bassal2012hebrew, cotter2015sociolinguistics}. These are mostly targeting academics and language learners. We consulted many of these resources as part of developing our annotation guidelines.



\paragraph{Dialectal Corpora}
We can group DA corpora based on the degree of richness in their annotations.
%
Some noteworthy examples of unannotated or lightly annotated corpora of relevance include the MADAR Corpus \cite{Bouamor:2018:madar}, comprising 2K parallel sentences spread across 25 dialects of Arabic, including PAL (Jerusalem variety) and the NADI corpus for nuanced dialect identification \cite{abdulmageed2021nadi}. The Shami Corpus \cite{abu-kwaik-etal-2018-shami} includes 21K PAL sentences, and the Parallel Arabic Dialect Corpus (PADIC) contains 6.4K PAL sentences \cite{Meftouh:2015:machine}. In the spirit of genre diversification and wider coverage across dialects, \newcite{el-haj-2020-habibi}  introduced the Habibi Corpus for song lyrics, which comprises songs from many Arab countries including all Levantine Arab countries.

Public and freely available morphologically annotated corpora are scarce for DA and often do not agree on annotation guidelines. A notable annotated dataset for PAL is the Curras corpus \cite{Jarrar:2016:curras}, a 56K-token morphologically annotated corpus. 
Other annotated  Levantine dialect efforts include the Jordan Comprehensive Contemporary Arabic Corpus (JCCA)
\cite{Sawalha:2019:construction}, the Jordanian and Syrian corpora by \newcite{alshargi:2019:morphologically}, and the
Baladi corpus of Lebanese Arabic \cite{alhaff-EtAl:2022:LREC}.


We consulted some of the public corpora as part of the development of Maknuune. However, most of the above datasets are based on web scrapes, which limits the amount of actual lemma coverage that they could attain.

\paragraph{Dialectal Lexicons} 
Examples of machine-readable DA lexicons include the 36K-lemma lexicon used for the CALIMA EGY fully inflected morphological analyzer \cite{Habash:2012:morphological}, based on the CALLHOME Egypt lexicon \cite{Gadalla:1997:callhome}, and the 51K-lemma Egyptian Arabic Tharwa lexicon  \cite{Diab:2014:tharwa}, which provides some morphological annotations.

The \textit{Palestinian Colloquial Arabic Vocabulary} comprises 4.5K entries including expressions \cite{younis2021palestinian}, and  the MADAR Lexicon contains 2.7K entries dedicated to the Jerusalem variety of PAL, including lemmas, phonological transcriptions, and glosses in MSA, English and French \cite{Bouamor:2018:madar}.

In addition to the above there are a number of dictionaries for Levantine Arabic variants, e.g., 
\newcite{elihai2004olive} (9K entries and 17K phrases for PAL), \newcite{moin2019etymological} (for PAL), \newcite{freiha:1973:dictionary} (ca. 5K entries for Lebanese Arabic),
and \newcite{stowasser2004dictionary} (15K entries for Syrian Arabic).
These resources include  base lemma forms, occasional plural forms, verb aspect inflections, and expressions; however,
none of them are available in a machine-readable format, to the best of our knowledge.

%
The lexicon presented in this work strives to be a large-scale and open resource with rich entries covering  phonology, morphology, and lexical expressions, and with a wide-ranging coverage of PAL sub-dialects. The lexicon may never be complete, but by making it open to sharing and contribution, we hope it will become central and useful to NLP researchers and developers, as well as to linguists working on Arabic and its dialects.

\section{Linguistic Facts}
\label{lingfacts}
In this section we present some general linguistic facts about PAL and highlight specific challenging phenomena that motivated many of our annotation decisions.

\subsection{Phonology and Orthography}
Like all other DA, and unlike MSA, PAL has no standard orthography rules \cite{Jarrar:2016:curras,Habash:2018:unified}.  In practice, PAL is primarily written in Arabic script, and to a lesser extent in Arabizi style romanization \cite{Darwish:2014:arabizi}. Some of the variations in the written form reflect the words' phonology, morphology, and/or etymological connections to MSA.  Orthogonal and detrimental to the orthography challenge, PAL has a high degree of variability within it sub-dialects in phonological terms. We highlight some below, noting that some also exist in other DA.

\paragraph{Consonantal Variables} 
A number of PAL consonants vary widely within sub-dialects. 
For example, the  voiceless velar stop \caphi{k} is affricated to the palatal  \caphi{tsh} in many PAL rural varieties \citep{herzallah1990aspects}, e.g., \<كَيف> 
{\it kayf} `how' appears as \caphi{k ee f} (urban) or \caphi{tsh ee f} (rural).\footnote{Arabic orthographic transliteration is presented in the HSB Scheme (italics) \cite{Habash:2007:arabic-transliteration}. Arabic script orthography is presented in the CODA* scheme, and Arabic phonology is presented in the CAPHI scheme (between /../) \cite{Habash:2018:unified}.}
Similarly, the MSA voiceless uvular stop \caphi{q} in the word \<قَلْب> 
{\it qal.b}
`heart'  is realized either as glottal stop \caphi{2 a l b} in urban dialects, as a voiceless velar stop \caphi{k a l b} in rural dialects, or a voiced velar stop \caphi{g a l b} in Bedouin dialects \citep{herzallah1990aspects}. 
It should be noted that there are some exceptions that do not conform to the above generalizations. For example, in Beit Fajjar,\footnote{A Palestinian town located 8 kilometers south of Bethlehem in the West Bank.} the word \<قَهْوَة> 
{\it qah.wa{\TAMARBUTA}}
`coffee' typically varying elsewhere  as \caphi{\{2,q,g,k\} a h w e} is realized as \caphi{tsh~h~ee~w~a}.
Moreover, some words do not have varying pronunciations such as \<عْقَال> 
{\it {\AYN}.qaAl}
\caphi{3~g~aa~l} `Egal headband'.


\paragraph{Monophthongization} 
Some PAL diphthongs shift to different monophthongs in different locations. 
For example the \caphi{a y} diphthong in \<شَيخ>
{\it {\SHIN}ayx} \caphi{sh~a~y~kh} `Sheikh' shifts often to \caphi{ee} (\caphi{sh ee kh}), but also to \caphi{ii} (\caphi{sh ii kh}).\footnote{In the Palestinian village of Ramadin, near Hebron in the West Bank.}
Following the CODA*  guidelines for diacritizing DA \cite{Habash:2018:unified}, we spell the \caphi{oo} and \caphi{ee} sounds using 
\<ىَو>~{\it aw}
and \<ىَي>~{\it ay} 
(without a \textit{sukun} on the \<و> \textit{w} or \<ي> 
\textit{y}), respectively, e.g.,
\<كَوم> \textit{kawm} \caphi{k oo m} `pile' and \<بَيت>
\textit{bayt} \caphi{b ee t} `house'.



\paragraph{Metathesis} 
In some rural dialects in villages near Tulkarem, Jenin and Ramallah, there are words with consonant pairs within a syllable that appear in a different order than is the norm in PAL, e.g., a word like \<كَهْرَبَا> 
{\it kah.rabaA} \caphi{k a h r a b a} `electricity' realizes as \caphi{k a r h a b a}.

\paragraph{Epenthesis}
PAL exhibits systematic epenthesis of the \caphi{i} or \caphi{u} sounds producing paired word alternations
such as \caphi{b a 3 d} and \caphi{b a 3 i d} for \<بعد> `still;after'
or
\caphi{kh u b z} and \caphi{kh u b u z} or \caphi{kh~u~b~i~z} (in different sub-dialects) for \<خبز> `bread'.
We opted to use the fully epenthesized forms in the lexicon, i.e., 
\<بَعِد>
\textit{ba{\AYN}id},
\<خُبُز>
\textit{xubuz},
and
\<خُبِز>
\textit{xubiz}, for the above mentioned examples.

\subsection{Morphology}

Like other DA, PAL has a complex morphology employing templatic and concatenative morphemes, and including a  rich set of morphological features: gender, number, person, state, aspect, in addition to numerous clitics.  We highlight some specific morphological phenomena that we needed to handle.

\paragraph{Ta Marbuta}
The so-called feminine singular suffix morpheme, or Ta Marbuta (\<ة> \TAMARBUTA), is a morpheme that can be used to mark feminine singular nominals, but that also appears with masculine singular and plural nominals.
Morphophonemically, it has a number of forms in PAL that vary contextually. 
First, in some PAL sub-dialects, the Ta Marbuta is pronounced as \caphi{a} when preceded by an emphatic consonant,  velars, and pharyngeal fricatives, e.g., 
\<بَطَّة>
{\it baT{\SHADDA}a{\TAMARBUTA}}
\caphi{b a t. t. a}
`duck'; otherwise it realizes as \caphi{e}, e.g., \<بِسِّة>
{\it bis{\SHADDA}i{\TAMARBUTA}}
\caphi{b i s s e}. 
In some northern PAL dialects, the \caphi{e} variant appears as \caphi{i}; and in some southern PAL dialects, the distinction is gone and all Ta Marbutas are pronounced \caphi{a}.
Second, the Ta Marbuta turns into its allomorph \caphi{i t} in {\it Idafa} constructions, e.g., \caphi{b i s s i t} `the/a cat of'. 
Finally, for some active participle deverbal nouns, the Ta Marbuta realizes as \caphi{aa} or \caphi{ii t} when followed by a pronominal object clitic, e.g., \<كَاتْبَاه>
{\it kaAt.baAh} \caphi{k aa t b aa (h)} or \<كَاتْبِيْتُه>
{\it kaAt.biy.tuh} or \caphi{k~a~t~b~ii~t~u~(h)} `she wrote it'.




\paragraph{Complex Plural Forms}
Besides the common use of broken plural (templatic plural) in DA, we encountered cases of {\it blocked} plurals where a typical sound plural or templatic plural is not generated because another word form is used in its place \citep{aronoff1976word}. One example from Ramadin, is the plural form  of 
 the word
\<عَيِّل>
{\it {\AYN}ay{\SHADDA}il} 
\caphi{3~a~y~y~i~l} `child [lit. dependent]', which is blocked by the word form \<ضْعُوف> 
{\it D.{\AYN}uwf} 
\caphi{dh.~3~uu~f} `children [lit. weaklings]'.

\subsection{Syntax}

Previous research on Arabic dialects reveals that the syntactic differences between these dialects
are considered to be minor compared to the morphological ones \citep{Brustad:2000:syntax}. 
%
%
One particular challenging phenomenon we encountered is a class of nouns used in adjectival constructions, but violating noun-adjective agreement rules, which involve gender, number and rationality \cite{Alkuhlani:2011:corpus}.  For instance, the word \<خِيخَة>
{\it xiyxa{\TAMARBUTA}} \caphi{kh~ii~kh~a} `weak/lame' does not typically agree with the nouns it modifies unlike a normal adjective such \<كْبِير>
{\it k.biyr} \caphi{k b ii r} `old [human]/large [nonhuman]'.  
So, the words
\<سِيَّارَة>
{\it siy{\SHADDA}aAra{\TAMARBUTA}} `car [f.s.]', 
\<عُرُس> {\it {\AYN}urus} `wedding [m.s.]', 
and \<نَاس> {\it naAs} `people [m.p]' can all be modified by \<خِيخَة>
{\it xiyxa{\TAMARBUTA}}; however, they need three different forms of \<كْبِير>
{\it k.biyr}: 
\<كْبِيرِة>
{\it k.biyri{\TAMARBUTA}},
\<كْبِير>
{\it k.biyr}, and
\<كْبَار>
{\it k.baAr}, respectively.
We mark the POS of such nominals as ADJ/NOUN in our lexicon, as it is a class that deserves further study.

\subsection{Figures of Speech and Multiword Expressions}

PAL has a rich culture of figures of speech and multiword expressions (compounds, collocations, etc.) that has not been well documented. We highlight some phenomena that we  cover in Maknuune.

\paragraph{Collocations}
As part of working on Maknuune, we encountered numeorus collocations (words that tend to co-occur with certain words more often than they do  with others). For example, the verbs used for trimming off the tough ends of some vegetables vary based on the vegetable:
\<يْقَمِّع بَامْيِا> 
\caphi{y~Q~a~m~m~i~3 \# b~aa~m~y~e} `trim off the tough ends of okra', \<يْقَرِّم فَاصَولْيَا>
\caphi{y~q~a~r~r~i~m \# f~aa~s.~uu~l~y~a} `trim off the tough ends of green beans', \<يْعَكِّب عَكُّوب>
\caphi{y~3~a~k~k~i~b \# 3~a~k~k~uu~b} `remove the thorns from artichoke (Gundelia)', and  \<يْطَرْطِف ذُرَة> 
\caphi{y~t.~a~r~t.~i~f \# D~u~r~a} `cut the blossom ends of the maize stalks'.

\paragraph{Compounds}
We encountered many compositional and non-compositional compounds. Examples include  \<جَوَاز سَفَر> 
{\it jawaAz safar}
\caphi{J a w aa z \# s~a~f~a~r} `[lit. permission-of-travel, passport]', which is also used in MSA. Some words appear in many compounds with a wide range of meaning, e.g.,
the word \<بَيت> {\it bayt} `[lit. house]' appears in compounds referring to celebrations, funerals, bathrooms, and whether or not a family has children (see the examples in  Table~\ref{tab:phrases}).

\paragraph{Synecdoches}
It has  been widely observed that PAL speakers use synecdoches\footnote{A figure of speech in which a term for a part of something is used to refer to the whole, or vice versa.} in their dialects \citep{seto1999distinguishing}.
Examples include the use of \<كَوم لَحِم> \caphi{k oo m \# l a 7 i m} `[lit. a pile of meat]', and \<كَبَابِيش>  \caphi{k~a~b~aa~b~ii sh} `[lit. plural of hair]' to mean `children'.


\paragraph{Euphemisms}
PAL speakers use many euphemistic expressions. For example, in some villages 
in Nablus, the expression \<يَوم تْهَنَّى> 
\caphi{y~oo~m~\# t h a n n a} `[lit. the day he felt happy]' to mean `the day he passed away'. 
In other areas in the West Bank, 
the phrase \<عَينُه كَرِيمِة> 
\caphi{3~ee~n~o~\# k~a~r~ii~m~e}
`[lit. his eye is generous]'
to mean `one-eyed'; and the phrase
\<بَيت خَالْتِي>
\caphi{b~ee~t~\# kh~aa~l~t~i}
`[lit. my aunt's house]' means 'prison'.


\section{Methodology}
\label{method}
In this section, we discuss the methodology  we adopted in data collection for Maknuune, as well as the guidelines we followed for creating the lexicon entries.

\subsection{Data Sources}
The current work spans over five years of effort, and a large number of volunteering informants, linguistics students, and citizen linguists (over 130 people).
The data was collected from many different sources.

First are \textbf{interviews} with (mostly but not entirely) elderly people  who live in rural areas such as villages and towns or in refugee camps in the West Bank.
The researchers went to the field and met with several people. 
They attended several social gatherings and participated in different events, e.g. weddings, funerals, field harvests, traditional cooking sessions, sewing, etc. They asked the language users several questions pertaining to the following themes: weddings, funerals, occupations, illnesses, cooking traditional dishes, plants, animals, myths, games, weather terms, tools and utensils, etc. They were particularly interested in documenting terms and expressions that are used mainly by the old generation. 

Secondly, to achieve the needed balance in the lexicon, the researchers consulted an in-house \textbf{balanced corpus}, that contains $\sim$40,000 words. The corpus comprises data that was transcribed from several recorded conversations that revolve around the same themes as above, written chats and texts, and some internet material (both written and spoken). Common words including verbs, adjectives, adverbs, and function words (e.g., prepositions, conjunctions, particles) were taken from the balanced corpus. At a later stage in the development of Maknuune, we consulted with the Curras Corpus \cite{Jarrar:2016:curras} to identify additional missing lemmas, with limited yield. We compare to Curras in terms of coverage in  Section~\ref{eval}. All of the above was also supplemented by methodical rounds of well-formedness checking to improve consistency across all fields, i.e., diacritization, transcription, root validity, etc.

Finally, in addition to the previous two methods, the researchers employed their \textbf{linguistic intuition} skills, knowledge of Palestinian Arabic (as native speakers) and the knowledge of the language users to provide additional word classes and multiword expressions that are associated with the existing lemmas.

It should be noted that whether an MSA lemma cognate of a PAL lemma (with similar or exact pronunciation, or meaning) exists was not considered a factor in including the PAL lemma in the lexicon.  We focused on creating a  representative sample of PAL including all its sub-dialects.

\begin{table*}[t!]
    \centering
\includegraphics[width=\linewidth]{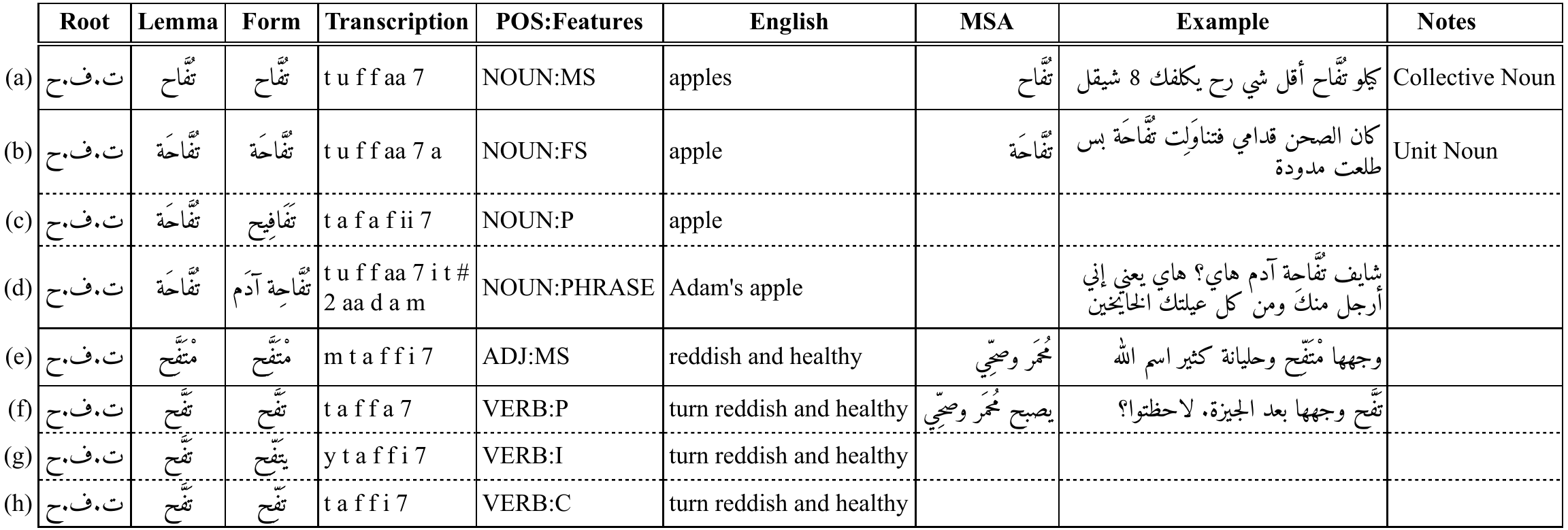}
    \caption{Eight entries from {\maknuune} that share the same root, and are paired with four distinct lemmas.}
    \label{tab:tf7}
\end{table*}

\subsection{Lexical Entries}

Each entry in the Maknuune lexicon consists of six required and three optional fields.
The six required fields are the \textbf{Root}, \textbf{Lemma}, \textbf{Form}, \textbf{Transcription},  \textbf{POS \& Features}, and \textbf{English Gloss}. The optional fields are the \textbf{MSA Gloss}, \textbf{Example} and \textbf{Notes}.
Figure~\ref{tab:tf7} presents an example of a number of entries coming from the same root.





\subsubsection{Root, Lemma, and Form}
The \textbf{Root}, \textbf{Lemma} and \textbf{Form} represent three degrees of morphological abstraction.
The \textbf{root} in Arabic in general is a templatic morpheme that interdigitates with a pattern or template to form a word stem that can then be inflected further. Roots are very abstract representations that broadly define the morphological family a word belongs to at the derivational and inflectional level. 
\textbf{Lemmas} on the other hand are abstractions of the inflectional space that is limited by variations in the morphological features of person, gender, number, aspect, etc. Lemmas are the central entries of the lexicon. 
\textbf{Forms} are base words (i.e., without clitics) that are inflected in a specific way. 
We follow the same general guidelines of determining lemmas as used in large Arabic morphological analyzers \cite{Graff:2009:standard,Habash:2012:morphological,Khalifa:2017:morphological}. There are of course some constructions that have grammaticalized into new lemmas, e.g., 
\<عَشَان> 
{\it {\AYN}a{\SHIN}aAn} can be treated as the noun  
\<شَان>
{\it {\SHIN}aAn} `situation;status' with a proclitic, or the subordinating conjunction meaning `because'.

For nouns and adjectives, we provide the lemma in the masculine singular form, unless it is a feminine form that does not vary in gender, in which case it is provided in the feminine singular. Very infrequently, some nouns only appear in plural form, which become their lemma, e.g. \<أَوَاعِي> {\it {\AHAMZAUP}awaA{\AYN}iy} \caphi{2~a~w~aa~3~i} `clothes'.  We do not list the sound plural and sound feminine inflections of nouns and adjectives. However, broken plurals and templatic feminine forms are provided and linked through the same lemma as the singular form.

For verbs, we provide the lemmas in the third masculine singular perfective form as is normally done in Arabic lexicography. We provide three forms linked to the lemma: the third masculine singular perfective, the third masculine singular imperfective, and the second person masculine imperative (command) forms.  These are provided for completeness to identify the basic verbal inflectional paradigm (albeit, not completely).

These three representations are provided in Arabic script.
Since PAL does not have an official standard orthography, we intentionally decided to follow the Conventional Orthography for Dialectal Arabic (CODA*) \cite{Habash:2018:unified}. In addition to being used in developing Curras \cite{Jarrar:2016:curras}, CODA* has been adopted  by a website for teaching PAL to non-native speakers.\footnote{\url{https://www.palestinianarabic.com/}}

\subsubsection{Transcription with CAPHI++}
One of CODA*'s limitations is that it abstracts over some of the phonological variations. As such, we follow the suggestions by \newcite{Habash:2018:unified} to use a phonological representation, CAPHI, to indicate the specific phonology of the entries.  CAPHI, which stands for Camel Phonetic Inventory is inspired by the  International Phonetic Alphabet (IPA)  and Arpabet \cite{Shoup:1980:phonological}, and is designed to only use characters directly accessible on the common keyboard to ease the job of annotators.

Owing to the phonological variations that are found in PAL, we extended CAPHI's symbol set with \textit{cover phonemes} that represent a number of possible interchangeable phones.  We call our extended set CAPHI++.  Table~\ref{tab:caphiplus} presents the new 9 symbols we introduced. All of these symbols are to be presented in upper case, while normal CAPHI symbols are in lower case. The new CAPHI++ symbols represent specific sets of mostly two variants in common use in different PAL sub-dialects.
For example, instead of including four entries for the word \<قَلَم> {\it qalam} 
(\caphi{q~a~l~a~m}, \caphi{k~a~l~a~m},  \caphi{2~a~l~a~m}, 
\caphi{g~a~l~a~m}),
we only provide one form (\caphi{Q~a~l~a~m}).
Exceptional usages that do not conform to the specific generalizations of the CAPHI++ cover symbols are listed independently, e.g., a second entry for the above example is provided for the Beit Fajjar pronunciation of \caphi{tsh~a~l~a~m}.

We acknowledge that the transcriptions provided may not represent the full breadth of PAL sub-dialects.  We make our resource open so that additional forms and variants can be added in the future, as needed.


\begin{table}[t!]
    \centering
    \includegraphics[width=\linewidth]{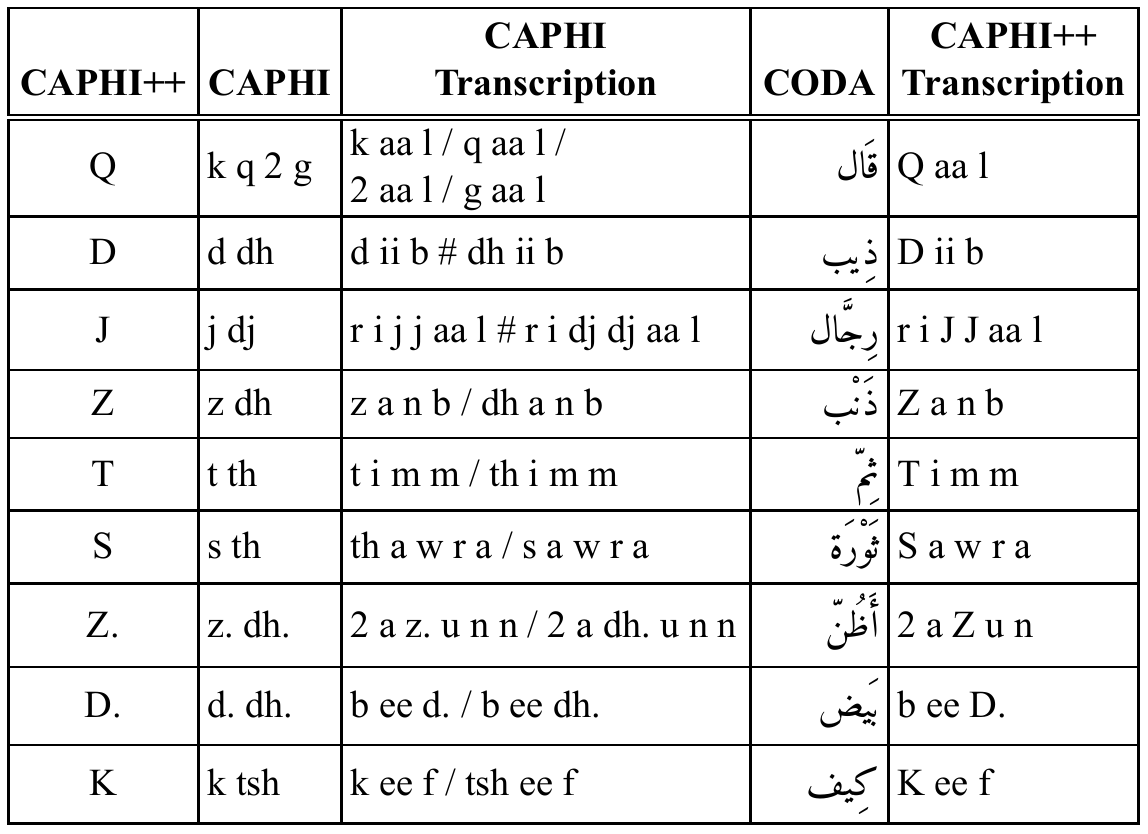}
    \caption{The CAPHI++ symbols set and its expanded CAPHI symbols, with examples.}
    \label{tab:caphiplus}
\end{table}

\subsubsection{POS and Features}

The analysis cell in every entry indicates the POS and features of the word form. 
We use 35 POS tags based on a combination of previously used POS tagsets in Arabic NLP \cite{Graff:2009:standard,Pasha:2014:madamira,Khalifa:2018:morphologically}.  Our closest relative is the tagset used by \cite{Khalifa:2018:morphologically} for work on Emirtai Arabic annotation. See the full list of POS tags in Table~\ref{tab:pos} in Appendix~\ref{pos-mapping}. 
However, we  extend their POS list with three tags: ADJ/NOUN (for adjectives with exceptional agreement), NOUN\_ACT (active participle deverbal noun), and NOUN\_PASS (passive participle deverbal noun).

For features, we use MS (masculine singular), FS (feminine singular), and P (plural) for nominals, 
and P (perfective), I (imperfective) and C (command) for third masculine singular verb forms only.

 




\subsubsection{Phrases} 
In addition to basic word forms, we overload the use of the form cells to list phrases (multiword expressions, collocations, and figures of speech) that are paired with the lemma. In such cases, the POS:Features cell is given the POS of the lemma, with the extension \textbf{PHRASE}, e.g., line (d) in Table~\ref{tab:tf7}, and 
Table~\ref{tab:phrases}.

\begin{table*}[th!]
    \centering
\includegraphics[width=\linewidth]{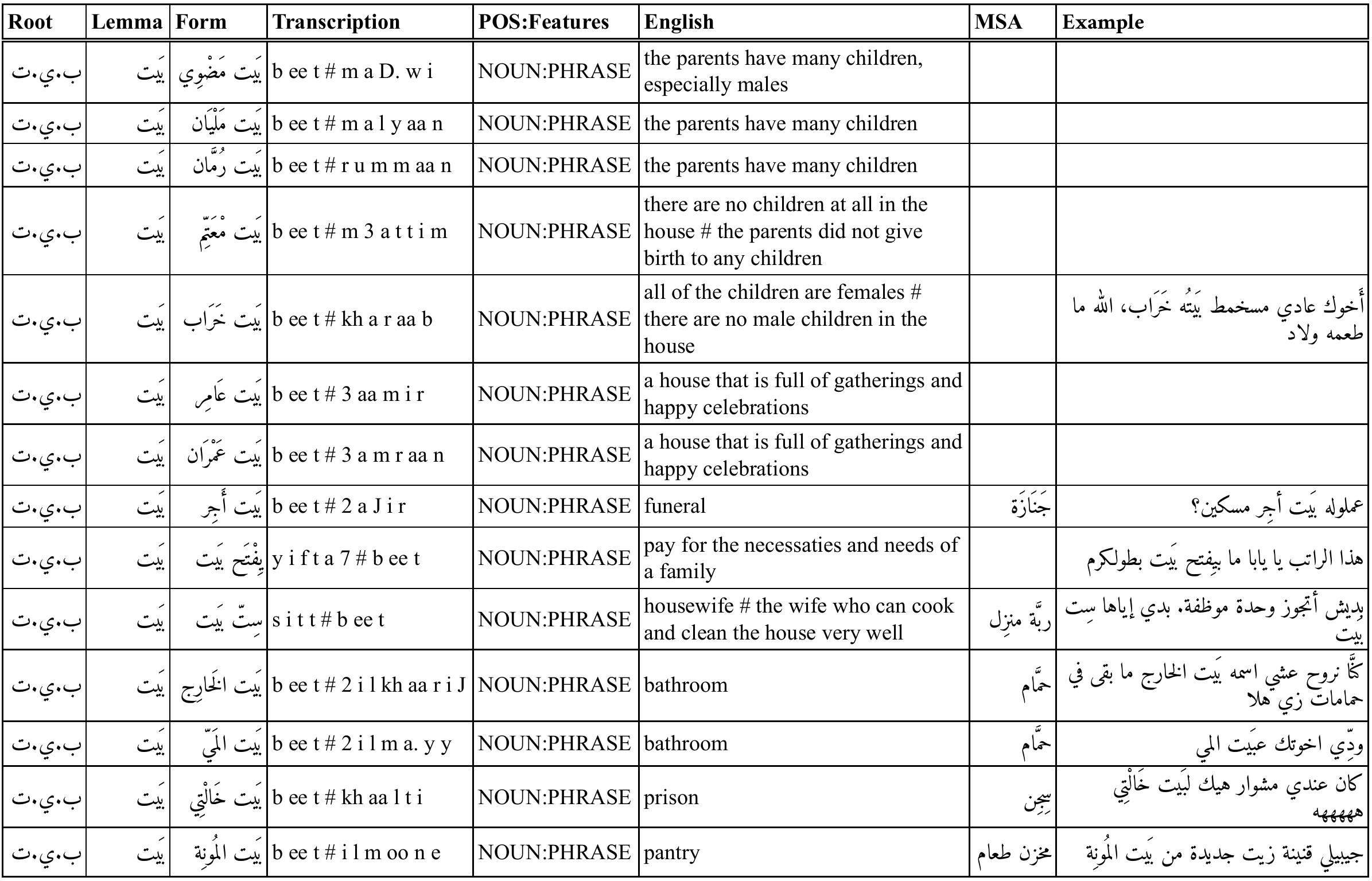}
    \caption{Examples of NC compounds in Maknuune for the lemma \<بَيت> `house'.}
    \label{tab:phrases}
\end{table*}

\subsubsection{Glosses, Examples and Notes}
We provided the English gloss equivalents of all the PAL words. The MSA gloss was provided for about a third of the entries at the time of writing. 
In cases where no single word in MSA or English can encode a culturally specific concept, the annotators translated the whole situation/concept. 
For example, in Ramadin, there are two words for `baby camel' depending on its age: \<ذَلُول> 
{\it {\DHA}aluwl} \caphi{dh~a~l~uu~l}, `barely a few days old'  and
\<حْوَيِّر>
{\it H.way{\SHADDA}ir} \caphi{7~w~a~y~y~i~r} `around 14-15 months old'. 
Another complex example is the word \<تَلْجِيم> {\it tal.jiym} \caphi{t a l J ii m} `[lit. harnessing or bridling]' which can refer also to `reciting some verses from the Quran (Surat Al-Takweer, Ayat Al-Kursi or Surat Al-Hashr) on a razor or a thread and closing the razor or tying the thread and leaving them aside until a lost or missing riding animal has returned home.' 

Finally, we provide usage examples for some entries, as well as grammatical or collection notes.  Notes vary in type from {\it Collective Noun} and {\it Collected near Nablus}, to {\it Vulgar}.




\section{Coverage Evaluation}
\label{eval}
We approximate the coverage of our lexicon by comparing it with the {\curras} corpus \cite{Jarrar:2016:curras}, the largest resource available for PAL.\footnote{\citet{alhaff-EtAl:2022:LREC} describe a revised version of that corpus, but it was not made available at the time of writing.} Since \curras is a corpus and our resource is a lexicon, the analysis is carried out in such a way to account for that difference. 
We  present next some high-level corpus statistics and then a detailed comparison between \maknuune and \curras.
Then, we  provide some comparison between \maknuune and the lexicons of two morphological analyzers for MSA and EGY.

\begin{table}[t]
    \centering
    \includegraphics[width=1\linewidth]{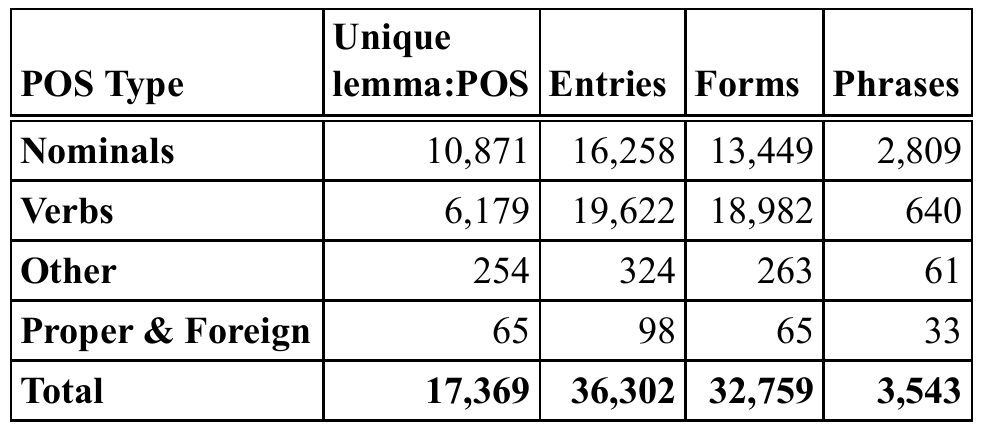}
    \caption{POS type and entry  statistics in \maknuune.}
    \label{fig:stats-maknuune}
\end{table}

\begin{table}[t]
    \centering
\includegraphics[width=1\linewidth]{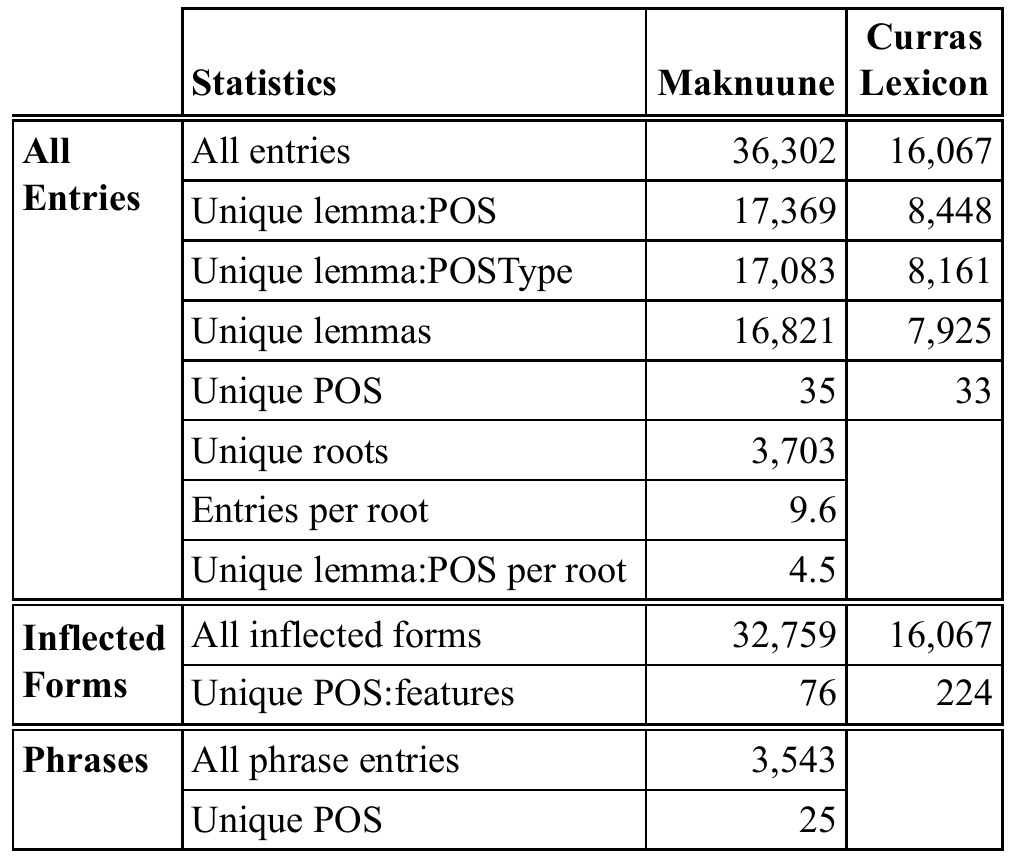}
    \caption{Side-by-side view of the statistics of both \maknuune and the lexicon extracted from \curras.}
    \label{fig:stats-comp}
\end{table}

\subsection{Maknuune \& Curras Statistics}
\paragraph{Maknuune POS Types}
Table \ref{fig:stats-maknuune} shows some basic statistics about \maknuune, dividing entries across four basic POS types (see Table~\ref{tab:pos}).
\maknuune has about three times more verb entries than verb lemmas, reflecting the fact that almost each verb appears in all three aspects (perfective, imperfective, and command) in third person masculine singular form. Similarly for nominals (nouns, adjectives, etc.), the ratio of 1.2 forms per lemma reflects the inclusion of plural entries for many nominals. 
Phrasal entries account for 10\% of all Maknuune entries, and close to three quarters of them are associated with nominals (63\% of all lemmas). 

\paragraph{The Curras Lexicon}

In order to compare \maknuune with \curras, we  extract a lexicon, henceforth Curras Lexicon, out of the Curras corpus by uniquing its entries based on lemma, inflected form, POS, and grammatical features (for \curras, aspect, person, gender, and number). 
We compare the Curras Lexicon to \maknuune in Table~\ref{fig:stats-comp}.

Firstly, Curras does not include  roots; and although it is a corpus, it does not identify phrases in the way Maknuune does. As such, we do not compare them in those terms in Table~\ref{fig:stats-comp}.

Secondly, by virtue of being a lexicon, \maknuune possesses more unique lemmas, weighing in at 17,369 lemmas taking POS into account (lemma:POS), while the total number of inflected forms is at 32,759, both of which are about 50\% more than in the Curras Lexicon. This clearly showcases \maknuune's richness in terms that go beyond the day-to-day language that one sees frequently in corpora like \curras. In contrast, \curras being a corpus, its extracted lexicon showcases a greater inflectional coverage with 224 unique word analyses as opposed to 76 for \maknuune. 

Finally, as inferable from the difference between the number of unique lemmas and lemma:POS, 548 lemmas are associated to more than one POS in \maknuune.

\subsection{Corpus Coverage Analysis}
\label{corpus-coverage-analysis}
In the interest of estimating how well our lexicon would fare with real-world data, we perform an analysis between the \curras and \maknuune lemmas, to see how many of the \curras lemmas \maknuune actually covers. From an initial investigation, we note that there are numerous minor differences that need to be normalized to ensure a more meaningful evaluation.
As such, we first pre-process all lemmas (in both lexicons) by stripping the \<سكون> {\it sukun} diacritic, stripping all the \<فتحة> diacritics that appear before a \<ا>~\textit{A},
converting the \<همزة وصل> \<ٱ>~\textit{Ä} to \<ا>~\textit{A}, and stripping the \<كسرة> (\textit{i}) and \<فتحة> (\textit{a}) diacritics if they appear before \<ة>~\textit{\TAMARBUTA}. We then compare all the annotated lemma:POSType
in \curras (56,004 tokens and 8,315 normalized types) to the lemmas in Maknuune.

We exclude 12,673 (23\%) of the tokens pertaining to punctuation, digits and proper noun POS, none of which were especially targeted by \maknuune. Of the remaining 43,331 entries, 49\% have exact match in \maknuune. We sample 10\% of the unique entries with no exact match (433 types and 1,965 tokens), and manually annotate them for their mismatch class.  We found that 74\% of all the sampled types (80\% in tokens) are actually present in \maknuune, but with slight differences in orthography mainly in the presence or absence of diacritics but also some spelling conventions. For about 20\% of sampled types (17\% in tokens), the lemma type is not one that we targeted such as foreign words and proper nouns that are differently labeled in \curras, or MSA words. Finally, 6\% of sampled types (3\% in tokens) are entries that are admittedly missing in \maknuune and can be added.

This suggests that we have very good coverage although the annotation errors and differences make it less obvious to see. A simple projected estimate assuming that our 10\% sample is representative would suggest that \maknuune's coverage of \curras' lexical terms (other than proper nouns and punctuation) is close to 94\% (97\% in token space); however a full detailed classification would be needed to confirm this projection. 

\subsection{Overlap with MSA and EGY}
In this section we conduct an evaluation similar to the one carried out in Section \ref{corpus-coverage-analysis} but with an MSA lexicon (Calima$_{MSA}$), and an Egyptian Arabic lexicon (Calima$_{EGY}$).\footnote{For MSA, we compared with the \texttt{calima-msa-s31\_0.4.2.utf8.db} version \cite{Taji:2018:arabic-morphological} based on SAMA \cite{Graff:2009:standard} and for EGY we only compared to the {\tt calima-egy-c044\_0.2.0.utf8.db} based on \newcite{Habash:2012:morphological}. For EGY, only {\tt CALIMA} analyses entries are selected.}
The analysis reveals that 44\% of \maknuune overlaps with Calima$_{MSA}$ at the lemma:POSType level (63\% if all entries are dediacritized),\footnote{The \textit{shadda} ({\SHADDA}) is not included in dediacritization.}
and that 49\% of \maknuune overlaps similarly with Calima$_{EGY}$ (75\% dediacritized). 
Taking into account that {\maknuune} spelling follows the CODA* guidelines,
the analysis suggests that the 37\% of {\maknuune}  lemma:POSTypes, which do not exist in the MSA lexicon we used, are heavily dialectal. The overlap with EGY is predictably higher, and the 25\% of Maknuune lemma:POSTypes (dediacritized) not existing in EGY highlights the differences between the two dialects despite their many similarities.

\subsection{Observations on Lexical Richness and Diversity}
The quantitative analyses we presented above allow us to see the big picture in terms of lexical richness and diversity in {\maknuune} and its complementarity to existing resources. However, we acknowledge that such an approach misses a lot of details that are collapsed or lost when ignoring subtle differences in semantics, phonology and morphology.

We first point at homonyms showing semantic changes and spread, such as  \<آوَى> 
/2 aa w a/ which is  `thread a needle' in PAL and ‘shelter sb’ in both MSA and PAL,
\<بَطّ> \caphi{b a t. t.} which means `very small olives that people find hard to pick' in some villages in Palestine and `ducks' in both MSA and PAL, and \<آخرة>
\caphi{2 aa kh r e} which means `desserts' in Nablus and `the Day of the Judgment' in both MSA and PAL, albeit with a different pronunciation. Clearly, additional entries are needed to mark these difference.

Furthermore, the majority of the entries in \maknuune are actually pronounced differently from MSA even if spelled the same without diacritics and thus warrant entries of their own, with clear phonological specifications.

Finally, if we consider morphology (which is not modeled here per se), many PAL lemmas that have MSA lemma cognates are actually inflected differently, e.g.,
\<مَدّ>
{\it mad{\SHADDA}} `extend;stretch'
(in PAL and MSA),
has different inflections for some parts of the paradigm: the 2nd person masculine plural is
\<مَدَّيتوا> {\it mad{\SHADDA}aytuwA} in PAL and
\<مَدَدْتُم> {\it madad.tum} in MSA.
Hence, each lemma in our lexicon heads a morphological paradigm which differs from its MSA counterpart.

\section{Conclusion and Future Work}
We presented Maknuune, a large open lexicon for the Palestinian Arabic dialect. Maknuune has 
over 36K entries from 17K lemmas, and 3.7K roots.  
All entries include Arabic diacritized orthography, phonological transcription and English glosses. 
Some entries are enriched with additional information such as  broken plural and templatic feminine forms, associated phrases and collocations, Standard Arabic glosses, and  examples or notes on grammar, usage, or location of collected entry.

In the future, we plan to continue to expand Maknuune to cover more PAL sub-dialects, more entries, and richer annotations, in particular for locations of usage, and morpholexical features such as rationality. We hope that by making it public, more researchers and citizen linguists will help enrich it and correct anything missing in it.

We also plan to make use of Maknuune as part of the development of larger resources and tools for Arabic NLP. The phonological transcriptions can be helpful for work in speech recognition and the morphological information for developing morphological analyzers and POS taggers. 
Furthermore, we plan to utilize Maknuune to develop pedagogical applications to help teach PAL to non-Arabic speakers and to children of Palestinians in the diaspora. 

\section*{Acknowledgments}
We would like to thank Prof. Jihad Hamdan, Muhammed Abu Odeh, Adnan Abu Shamma, Issra Ghazzawi and Kazem Abu-Khalaf for the helpful discussions.

\bibliography{camel-bib-v2,extra}

\begin{thebibliography}{38}
\expandafter\ifx\csname natexlab\endcsname\relax\def\natexlab#1{#1}\fi

\bibitem[{Abdul-Mageed et~al.(2021)Abdul-Mageed, Zhang, Elmadany, Bouamor, and
  Habash}]{abdulmageed2021nadi}
Muhammad Abdul-Mageed, Chiyu Zhang, AbdelRahim Elmadany, Houda Bouamor, and
  Nizar Habash. 2021.
\newblock \href {https://aclanthology.org/2021.wanlp-1.28} {{NADI} 2021: The
  second nuanced {A}rabic dialect identification shared task}.
\newblock In \emph{Proceedings of the Sixth Arabic Natural Language Processing
  Workshop}, pages 244--259, Kyiv, Ukraine (Virtual). Association for
  Computational Linguistics.

\bibitem[{Abu~Kwaik et~al.(2018)Abu~Kwaik, Saad, Chatzikyriakidis, and
  Dobnik}]{abu-kwaik-etal-2018-shami}
Kathrein Abu~Kwaik, Motaz Saad, Stergios Chatzikyriakidis, and Simon Dobnik.
  2018.
\newblock \href {https://aclanthology.org/L18-1576} {{S}hami: A corpus of
  {L}evantine {A}rabic dialects}.
\newblock In \emph{Proceedings of the Eleventh International Conference on
  Language Resources and Evaluation ({LREC} 2018)}, Miyazaki, Japan. European
  Language Resources Association (ELRA).

\bibitem[{Al-Haff et~al.(2022)Al-Haff, Jarrar, Hammouda, and
  Zaraket}]{alhaff-EtAl:2022:LREC}
Karim Al-Haff, Mustafa Jarrar, Tymaa Hammouda, and Fadi Zaraket. 2022.
\newblock \href {https://aclanthology.org/2022.lrec-1.82} {{Curras + Baladi:
  Towards a Levantine Corpus}}.
\newblock In \emph{Proceedings of the Language Resources and Evaluation
  Conference}, pages 769--778, Marseille, France. European Language Resources
  Association.

\bibitem[{Alkuhlani and Habash(2011)}]{Alkuhlani:2011:corpus}
Sarah Alkuhlani and Nizar Habash. 2011.
\newblock {A Corpus for Modeling Morpho-Syntactic Agreement in {{A}rabic:}
  Gender, Number and Rationality}.
\newblock In \emph{Proceedings of the Conference of the Association for
  Computational Linguistics (ACL)}, Portland, Oregon, USA.

\bibitem[{Alshargi et~al.(2019)Alshargi, Dibas, Alkhereyf, Faraj, Abdulkareem,
  Yagi, Kacha, Habash, and Rambow}]{alshargi:2019:morphologically}
Faisal Alshargi, Shahd Dibas, Sakhar Alkhereyf, Reem Faraj, Basmah Abdulkareem,
  Sane Yagi, Ouafaa Kacha, Nizar Habash, and Owen Rambow. 2019.
\newblock \href {https://doi.org/10.18653/v1/W19-4615} {Morphologically
  annotated corpora for seven {A}rabic dialects: Taizi, sanaani, najdi,
  jordanian, syrian, iraqi and {M}oroccan}.
\newblock In \emph{Proceedings of the Fourth Arabic Natural Language Processing
  Workshop}, pages 137--147, Florence, Italy. Association for Computational
  Linguistics.

\bibitem[{Alyafeai et~al.(2022)Alyafeai, Masoud, Ghaleb, and
  Al-shaibani}]{alyafeai2022masader}
Zaid Alyafeai, Maraim Masoud, Mustafa Ghaleb, and Maged~S. Al-shaibani. 2022.
\newblock \href {https://arbml.github.io/masader/} {Masader: Metadata sourcing
  for {A}rabic text and speech data resources}.
\newblock In \emph{Proceedings of the Language Resources and Evaluation
  Conference}, Marseille, France.

\bibitem[{Aronoff(1976)}]{aronoff1976word}
Mark Aronoff. 1976.
\newblock Word formation in generative grammar.
\newblock \emph{Linguistic Inquiry, Monograph one, The MIT press}.

\bibitem[{Bassal(2012)}]{bassal2012hebrew}
Ibrahim Bassal. 2012.
\newblock {Hebrew and Aramaic Substrata in Spoken Palestinian Arabic}.
\newblock \emph{Mediterranean Language Review}, 19:85--104.

\bibitem[{Bouamor et~al.(2018)Bouamor, Habash, Salameh, Zaghouani, Rambow,
  Abdulrahim, Obeid, Khalifa, Eryani, Erdmann, and
  Oflazer}]{Bouamor:2018:madar}
Houda Bouamor, Nizar Habash, Mohammad Salameh, Wajdi Zaghouani, Owen Rambow,
  Dana Abdulrahim, Ossama Obeid, Salam Khalifa, Fadhl Eryani, Alexander
  Erdmann, and Kemal Oflazer. 2018.
\newblock {The MADAR {A}rabic Dialect Corpus and Lexicon}.
\newblock In \emph{Proceedings of the Language Resources and Evaluation
  Conference (LREC)}, Miyazaki, Japan.

\bibitem[{Brustad(2000)}]{Brustad:2000:syntax}
Kristen Brustad. 2000.
\newblock \emph{The Syntax of Spoken {A}rabic: A Comparative Study of Moroccan,
  Egyptian, Syrian, and Kuwaiti Dialects}.
\newblock Georgetown University Press.

\bibitem[{Cotter and Horesh(2015)}]{cotter2015sociolinguistics}
William Cotter and Uri Horesh. 2015.
\newblock {Sociolinguistics of Palestinian Arabic}.
\newblock \emph{Encyclopedia of Arabic Language \& Linguistics}.

\bibitem[{Darwish(2014)}]{Darwish:2014:arabizi}
Kareem Darwish. 2014.
\newblock {Arabizi Detection and Conversion to {A}rabic}.
\newblock In \emph{Proceedings of the Workshop for {A}rabic Natural Language
  Processing (WANLP)}, pages 217--224, Doha, Qatar.

\bibitem[{Diab et~al.(2014)Diab, Al-Badrashiny, Aminian, Attia, Elfardy,
  Habash, Hawwari, Salloum, Dasigi, and Eskander}]{Diab:2014:tharwa}
Mona~T Diab, Mohamed Al-Badrashiny, Maryam Aminian, Mohammed Attia, Heba
  Elfardy, Nizar Habash, Abdelati Hawwari, Wael Salloum, Pradeep Dasigi, and
  Ramy Eskander. 2014.
\newblock {Tharwa: A Large Scale Dialectal {A}rabic-Standard {A}rabic-{E}nglish
  Lexicon}.
\newblock In \emph{Proceedings of the Language Resources and Evaluation
  Conference (LREC)}, pages 3782--3789, Reykjavik, Iceland.

\bibitem[{El-Haj(2020)}]{el-haj-2020-habibi}
Mahmoud El-Haj. 2020.
\newblock \href {https://aclanthology.org/2020.lrec-1.165} {Habibi - a multi
  dialect multi national {A}rabic song lyrics corpus}.
\newblock In \emph{Proceedings of the 12th Language Resources and Evaluation
  Conference}, pages 1318--1326, Marseille, France. European Language Resources
  Association.

\bibitem[{Elihai(2004)}]{elihai2004olive}
Yohanan Elihai. 2004.
\newblock \emph{{The olive tree dictionary: A transliterated dictionary of
  conversational Eastern Arabic (Palestinian)}}.
\newblock Minerva Jerusalem.

\bibitem[{Ferguson(1959)}]{Ferguson:1959:diglossia}
Charles~F Ferguson. 1959.
\newblock {Diglossia}.
\newblock \emph{Word}, 15(2):325--340.

\bibitem[{Freiha(1973)}]{freiha:1973:dictionary}
Anis Freiha. 1973.
\newblock \emph{Dictionary of Non-Classical Vocables in the Spoken Arabic of
  Lebanon}.
\newblock Librairie du Liban.

\bibitem[{Gadalla et~al.(1997)Gadalla, Kilany, Arram, Yacoub, El-Habashi,
  Shalaby, Karins, Rowson, MacIntyre, Kingsbury, Graff, and
  McLemore}]{Gadalla:1997:callhome}
Hassan Gadalla, Hanaa Kilany, Howaida Arram, Ashraf Yacoub, Alaa El-Habashi,
  Amr Shalaby, Krisjanis Karins, Everett Rowson, Robert MacIntyre, Paul
  Kingsbury, David Graff, and Cynthia McLemore. 1997.
\newblock {CALLHOME} {E}gyptian {A}rabic transcripts {LDC97T19}.
\newblock Web Download. Philadelphia: Linguistic Data Consortium.

\bibitem[{Graff et~al.(2009)Graff, Maamouri, Bouziri, Krouna, Kulick, and
  Buckwalter}]{Graff:2009:standard}
David Graff, Mohamed Maamouri, Basma Bouziri, Sondos Krouna, Seth Kulick, and
  Tim Buckwalter. 2009.
\newblock {Standard {A}rabic Morphological Analyzer (SAMA) Version 3.1}.
\newblock Linguistic Data Consortium LDC2009E73.

\bibitem[{Habash et~al.(2018)Habash, Eryani, Khalifa, Rambow, Abdulrahim,
  Erdmann, Faraj, Zaghouani, Bouamor, Zalmout, Hassan, shargi, Alkhereyf,
  Abdulkareem, Eskander, Salameh, and Saddiki}]{Habash:2018:unified}
Nizar Habash, Fadhl Eryani, Salam Khalifa, Owen Rambow, Dana Abdulrahim,
  Alexander Erdmann, Reem Faraj, Wajdi Zaghouani, Houda Bouamor, Nasser
  Zalmout, Sara Hassan, Faisal~Al shargi, Sakhar Alkhereyf, Basma Abdulkareem,
  Ramy Eskander, Mohammad Salameh, and Hind Saddiki. 2018.
\newblock Unified guidelines and resources for {A}rabic dialect orthography.
\newblock In \emph{Proceedings of the Language Resources and Evaluation
  Conference (LREC)}, Miyazaki, Japan.

\bibitem[{Habash et~al.(2012)Habash, Eskander, and
  Hawwari}]{Habash:2012:morphological}
Nizar Habash, Ramy Eskander, and Abdelati Hawwari. 2012.
\newblock A {M}orphological {A}nalyzer for {E}gyptian {A}rabic.
\newblock In \emph{Proceedings of the Workshop of the Special Interest Group on
  Computational Morphology and Phonology (SIGMORPHON)}, pages 1--9,
  Montr\'{e}al, Canada.

\bibitem[{Habash et~al.(2007)Habash, Soudi, and
  Buckwalter}]{Habash:2007:arabic-transliteration}
Nizar Habash, Abdelhadi Soudi, and Tim Buckwalter. 2007.
\newblock {On {{A}rabic} Transliteration}.
\newblock In A.~van~den Bosch and A.~Soudi, editors, \emph{{A}rabic
  Computational Morphology: Knowledge-based and Empirical Methods}, pages
  15--22. Springer, Netherlands.

\bibitem[{Halloun(2019)}]{moin2019etymological}
Moïn Halloun. 2019.
\newblock \emph{An etymological lexicon of foreign words in Palestinian
  Arabic~: Arabic-Arabic-English~: the influence of Greek, Pahlavi, Latin,
  Persian Syriac, Ottoman language and modern languages in the Palestinian
  dialect}.
\newblock Bethlehem: Bethlehem University, The Institute of Oral Cultural
  Heritage of the Palestinians.

\bibitem[{Herzallah(1990)}]{herzallah1990aspects}
Rukayyah~S Herzallah. 1990.
\newblock \emph{{Aspects of Palestinian Arabic phonology: A nonlinear
  approach}}.
\newblock Cornell University.

\bibitem[{Hopkins(1995)}]{hopkins1995sarar}
Simon Hopkins. 1995.
\newblock sar{\=a}r "pebbles" — {A Canaanite Substrate Word in Palestinian
  Arabic}.
\newblock \emph{Zeitschrift f{\"u}r arabische Linguistik}, (30):37--49.

\bibitem[{Jarrar et~al.(2016)Jarrar, Habash, Alrimawi, Akra, and
  Zalmout}]{Jarrar:2016:curras}
Mustafa Jarrar, Nizar Habash, Faeq Alrimawi, Diyam Akra, and Nasser Zalmout.
  2016.
\newblock {Curras: an annotated corpus for the Palestinian {A}rabic dialect}.
\newblock \emph{Language Resources and Evaluation}, pages 1--31.

\bibitem[{Khalifa et~al.(2018)Khalifa, Habash, Eryani, Obeid, Abdulrahim, and
  Kaabi}]{Khalifa:2018:morphologically}
Salam Khalifa, Nizar Habash, Fadhl Eryani, Ossama Obeid, Dana Abdulrahim, and
  Meera~Al Kaabi. 2018.
\newblock A morphologically annotated corpus of {E}mirati {A}rabic.
\newblock In \emph{Proceedings of the Language Resources and Evaluation
  Conference (LREC)}, Miyazaki, Japan.

\bibitem[{Khalifa et~al.(2017)Khalifa, Hassan, and
  Habash}]{Khalifa:2017:morphological}
Salam Khalifa, Sara Hassan, and Nizar Habash. 2017.
\newblock A morphological analyzer for {G}ulf {A}rabic verbs.
\newblock In \emph{Proceedings of the Workshop for {A}rabic Natural Language
  Processing (WANLP)}, Valencia, Spain.

\bibitem[{Meftouh et~al.(2015)Meftouh, Harrat, Jamoussi, Abbas, and
  Smaili}]{Meftouh:2015:machine}
Karima Meftouh, Salima Harrat, Salma Jamoussi, Mourad Abbas, and Kamel Smaili.
  2015.
\newblock Machine translation experiments on {PADIC}: A parallel {A}rabic
  dialect corpus.
\newblock In \emph{Proceedings of the Pacific Asia Conference on Language,
  Information and Computation}.

\bibitem[{Pasha et~al.(2014)Pasha, Al-Badrashiny, Diab, Kholy, Eskander,
  Habash, Pooleery, Rambow, and Roth}]{Pasha:2014:madamira}
Arfath Pasha, Mohamed Al-Badrashiny, Mona Diab, Ahmed~El Kholy, Ramy Eskander,
  Nizar Habash, Manoj Pooleery, Owen Rambow, and Ryan Roth. 2014.
\newblock {MADAMIRA}: A fast, comprehensive tool for morphological analysis and
  disambiguation of {A}rabic.
\newblock In \emph{Proceedings of the Language Resources and Evaluation
  Conference (LREC)}, pages 1094--1101, Reykjavik, Iceland.

\bibitem[{Rice and Sa'id(1979)}]{Rice:1979:eastern}
Frank Rice and Majed Sa'id. 1979.
\newblock \emph{Eastern {A}rabic}.
\newblock Georgetown University Press.

\bibitem[{Sawalha et~al.(2019)Sawalha, Alshargi, AlShdaifat, Yagi, and
  Qudah}]{Sawalha:2019:construction}
Majdi Sawalha, Faisal Alshargi, Abdallah AlShdaifat, Sane Yagi, and Mohammad~A.
  Qudah. 2019.
\newblock \href {https://doi.org/10.18653/v1/W19-4616} {Construction and
  annotation of the {J}ordan comprehensive contemporary {A}rabic corpus
  ({JCCA})}.
\newblock In \emph{Proceedings of the Fourth Arabic Natural Language Processing
  Workshop}, pages 148--157, Florence, Italy. Association for Computational
  Linguistics.

\bibitem[{Seto(1999)}]{seto1999distinguishing}
Ken-ichi Seto. 1999.
\newblock Distinguishing metonymy from synecdoche.
\newblock \emph{Metonymy in language and thought}, 4:91--120.

\bibitem[{Shoup(1980)}]{Shoup:1980:phonological}
June~E Shoup. 1980.
\newblock Phonological aspects of speech recognition.
\newblock \emph{Trends in Speech Recognition}, pages 125--138.

\bibitem[{Stowasser and Ani(2004)}]{stowasser2004dictionary}
K.~Stowasser and M.~Ani. 2004.
\newblock \href {https://books.google.com/books?id=dRScfbxJ45IC} {\emph{A
  Dictionary of Syrian Arabic: English-Arabic}}.
\newblock G - Reference, Information and Interdisciplinary Subjects Series.
  Georgetown University Press.

\bibitem[{Taji et~al.(2018)Taji, Khalifa, Obeid, Eryani, and
  Habash}]{Taji:2018:arabic-morphological}
Dima Taji, Salam Khalifa, Ossama Obeid, Fadhl Eryani, and Nizar Habash. 2018.
\newblock {A}n {A}rabic {M}orphological {A}nalyzer and {G}enerator with
  {C}opious {F}eatures.
\newblock In \emph{Proceedings of the Fifteenth Workshop on Computational
  Research in Phonetics, Phonology, and Morphology (SIGMORPHON)}, pages
  140--150.

\bibitem[{Talmon(2004)}]{talmon200419th}
Raphael Talmon. 2004.
\newblock {19th century Palestinian Arabic: the testimony of Western
  travellers}.
\newblock \emph{Jerusalem studies in Arabic and Islam}, (29):210--280.

\bibitem[{Younis and Aldrich(2021)}]{younis2021palestinian}
A.~Younis and M.~Aldrich. 2021.
\newblock \href {https://books.google.com/books?id=2d23zgEACAAJ}
  {\emph{Palestinian Colloquial Arabic Vocabulary}}.
\newblock Arabic Vocabulary. Lingualism.

\end{thebibliography}
\bibliographystyle{acl_natbib}

\appendix

\section{POS Type Mapping and Examples}
\label{pos-mapping}

\begin{table}[h!]
    \centering
\includegraphics[width=1\linewidth]{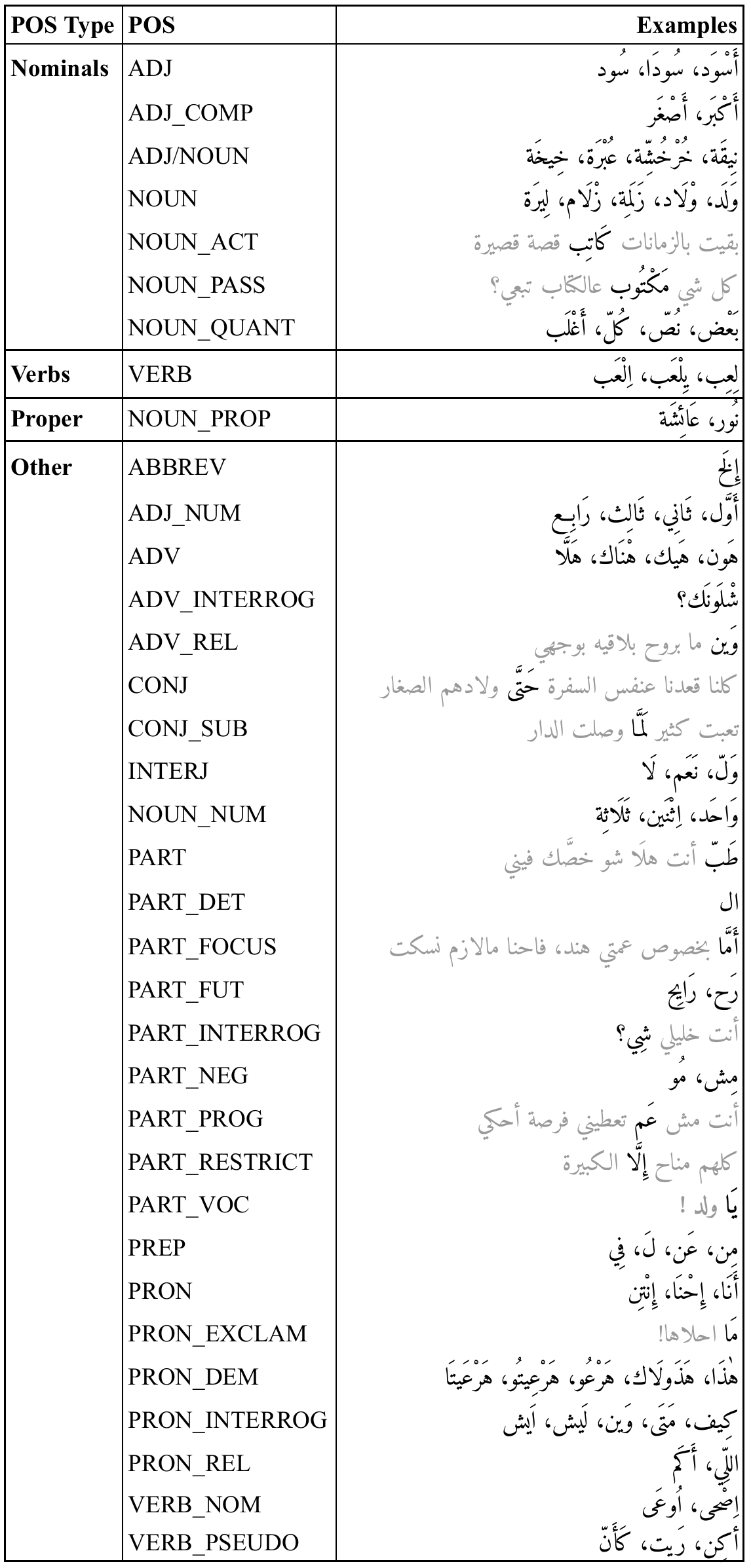}
    \caption{Mapping of part-of-speech (POS) types to POS tags used to annotate base words in Maknuune, and associated examples.}
    \label{tab:pos}
\end{table}

\end{document}